\documentclass[runningheads]{llncs}
\usepackage{graphicx}
\usepackage{amsmath}
\usepackage{amssymb}
\usepackage{algorithm}
\usepackage{algorithmic}
\usepackage{multirow}
\usepackage{booktabs}
\usepackage{wrapfig}
\usepackage{color}
\usepackage{array}
\usepackage{url}
\usepackage{threeparttable}

\begin{document}
\graphicspath{{Imgs/}}

\makeatletter
\renewcommand{\overset}[2]{\ensuremath{\mathop{\kern\z@\mbox{#2}}\limits^{\mbox{\scriptsize #1}}}}
\renewcommand{\underset}[2]{\ensuremath{\mathop{\kern\z@\mbox{#2}}\limits_{\mbox{\scriptsize #1}}}}
\makeatother

\title{Meta Corrupted Pixels Mining for Medical \\Image Segmentation}
\titlerunning{Meta Corrupted Pixels Mining}
\ifx\anonymous\undifined
    \author{Jixin Wang\inst{1}
    \and
    Sanping Zhou\inst{1} 
    \and
    Chaowei Fang\inst{2}
    \and
    Le Wang \inst{1} 
    \and
    Jinjun Wang\inst{1}\thanks{Corresponding author}}
    \authorrunning{J. X. Wang et al.}
    \institute{Institute of Artificial Intelligence and Robotics\\Xi’an Jiaotong University, Xi’an, Shaanxi, P.R. China \and
    School of Artificial Intelligence\\ Xidian University, Xi'an, Shaanxi, P.R. China}
\fi
%

%

%
\maketitle              
\begin{abstract}
Deep neural networks have achieved satisfactory performance in piles of medical image analysis tasks. However the training of deep neural network requires a large amount of samples with high-quality annotations. In medical image segmentation, it is very laborious and expensive to acquire precise pixel-level annotations. Aiming at training deep segmentation models on datasets with probably corrupted annotations, we propose a novel Meta Corrupted Pixels Mining~(MCPM) method based on a simple meta mask network. Our method is targeted at automatically estimate a weighting map to evaluate the importance of every pixel in the learning of segmentation network. The meta mask network which regards the loss value map of the predicted segmentation results as input, is capable of identifying out corrupted layers and allocating small weights to them. An alternative algorithm is adopted to train the segmentation network and the meta mask network, simultaneously. Extensive experimental results on LIDC-IDRI and LiTS datasets show that our method outperforms state-of-the-art approaches which are devised for coping with corrupted annotations.



\keywords{Meta Corrupted Pixels Mining \and Deep Neural Network \and Medical Image Segmentation }
\end{abstract}

\section{Introduction}
Recent years have witnessed the blooming of Deep Neural Networks (DNNs) in medical image analysis, including image segmentation, image registration, image reconstruction~\cite{Liu_Xu_Wu:2018}, and etc. Due to the powerful representation capability of DNN, significant progress has been achieved in medical image analysis. However, training a DNN usually requires a large number of high-quality labeled samples, which is hard to acquire in various applications. For example, it is very expensive to generate a precise segment of input image, because the pathological tissue needs to be marked by professional radiologists~\cite{Zhou_Wang_Zhang:2017,Zhou_Wang_Zhang:2016}. As a result, a question was naturally raised: How can we train a powerful segmentation network only using a small number of high-quality labeled samples?

\begin{figure}[t]
	\centering
	\includegraphics[height = 3.0cm, width = 9.0cm]{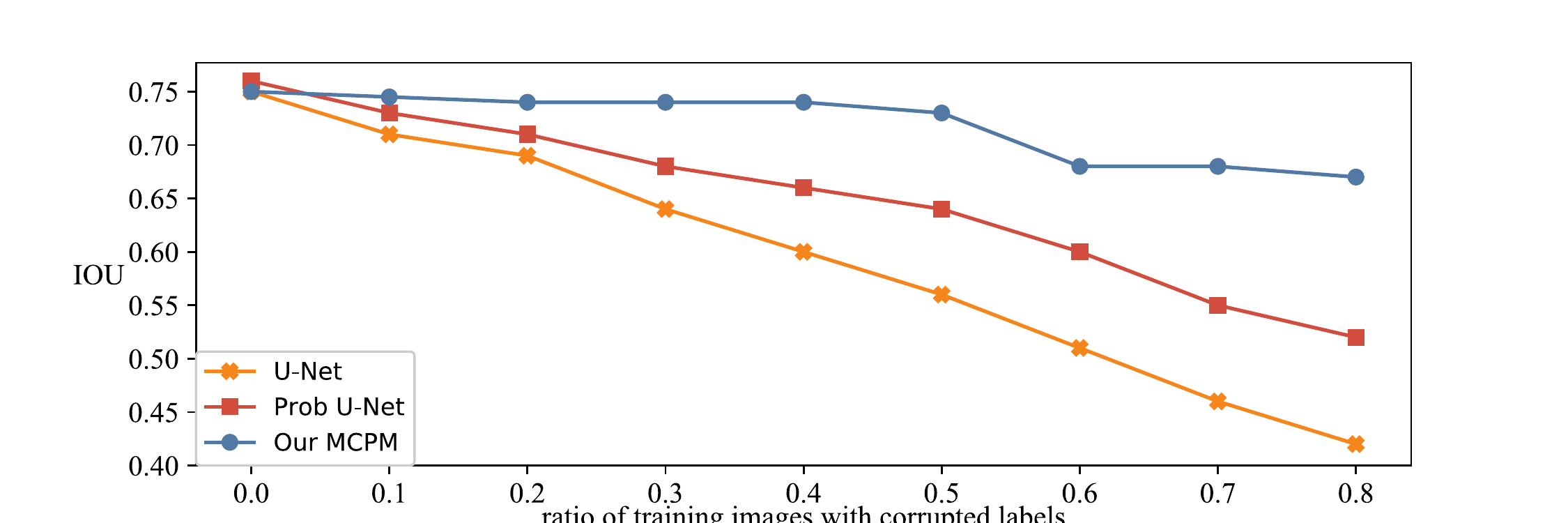}
	\caption{Corrupted labels effect network's performance}
	\label{fig_1}
\end{figure}

To address this situation, researchers have paid much attention to train DNNs in a semi-supervised manner. For example, Yang et al.~\cite{Yang_Zhang_Chen:2017} presented an active learning method for 2D biomedical image segmentation, which can improve segmentation accuracy through suggesting the most effective rather than all samples for labeling. In~\cite{Zhao_Yang_Zheng:2018}, Zhao et al. applied a modified Mask R-CNN to volumetric data for instance segmentation, and they used bounding boxes for all instances and voxel-wise labels for a small proportion of instances. Nie et al.~\cite{Nie_Gao_Wang:2018} proposed an attention based semi-supervised deep networks, which adopted the adversarial learning strategy to deal with the insufficient data problem in training complex networks. In practice, the success of these semi-supervised methods depends on mining a kind of knowledge which can be used to find out more accurate labels in the training process. However, most of the existing methods use a fixed prior knowledge to guide the pseudo label estimation. Therefore, they are very unstable when dealing with training samples with complex noise distributions. As shown in Fig.~\ref{fig_1}, the segmentation network's results are seriously affected when corrupted labels are taken as supervisory signals. This phenomenon reveals that mining corrupted labels is a critical issue in semi-supervised image segmentation.

In this paper, we design a novel Meta Corrupted Pixels Mining~(MCPM) method for medical image segmentation, which can alleviate the impacts of corrupted labels in the training process. To achieve this goal, we design a simple meta mask network to protect the training of the segmentation network from the influence of pixels with incorrect labels. Specifically, the meta mask network absorbs in the loss value map of the segmentation prediction as input, and estimate a weight map indicating the importance of every pixel in the training of the segmentation network. Once the meta mask network is learned, small weights are allocated to pixels with corrupted labels. Therefore influences from these pixels are weakened when updating the segmentation network. In the training process, we update the segmentation network and meta mask network in an alternate manner, which can learn a powerful segmentation network from images with corrupted labels. The main contributions of this work can be highlighted as follows:
\begin{itemize}
	\item We design a novel meta learning framework to mine pixels with corrupted labels during the process of training a segmentation network.
	\item Based on the fully convolutional structure, we build up a meta mask network which can automatically estimate pixel-wise importance factors for mitigating the influence of corrupted labels.
	\item Extensive experiments on both LIDC-IDRI and LiTS datasets indicate that our method achieves the state-of-the-art performance in medical image segmentation with incorrect labels.
\end{itemize}

\section{Related Works}
Because our method takes U-Net~\cite{unet} as segmentation network and applies the meta learning regime~\cite{metalearning} to mine pixels of corrupted labels, we briefly review a few existing works in terms of U-Net and meta learning in the following paragraphs.

\noindent \textbf{Methods based on U-Net}. This type of methods aim to design a powerful network structure, which can obtain accurate segmentation results at the output layer. In~\cite{unet}, Ronneberger et al. proposed a well-known U-shaped structure for 2D medical image segmentation, in which the low-level and high-level feature are recursively concatenated together from top to down, to improve segmentation results. Inspired by this idea, a number of variants have been introduced in the past few years. For example, Milletari et al.~\cite{vnet} extended the U-shaped structure into 3D version and built an objective function and adopted Dice coefficient maximisation to supervise the training process. In~\cite{probunet}, Kohl et al. proposed a generative segmentation model based on a combination of a U-Net, in which a conditional variational autoencoder is designed to produce an unlimited number of plausible hypotheses. Because its superior performance in medical image segmentation, we simply choose U-Net as our segmentation network. Then, we concentrate on designing a meta learning regime which can help learn a robust segmentation network from training samples with corrupted labels.

\noindent \textbf{Methods based on Meta Learning}. This kind of methods aim to learn a kind of knowledge which can be used to guide the training of the network for solving the target problem~\cite{metalearning}, which has a wide application in the few-shot learning community. For example, a number of methods, such as FWL~\cite{Dehghani_Mehrjou_Gouws:2017}. MentorNet~\cite{Jiang_Zhou_Leung:2018} used the concept of meta learning to learn an adaptive weighting function to make the training process more robust to noisy images. However, the meta learners used in these methods have complex forms and require complicated inputs, which are very hard to be implemented in the training process. To overcome this problem, Ren et al.~\cite{l2rw} proposed a novel meta learning algorithm which can learn an implicit function to assign weights to training samples based on their gradient directions. In~\cite{meta_weight_learning}, Shu et al. designed a meta weight network to lean an explicit function which can impose small weights to noisy samples, therefore the noisy samples will not severely affect the training process. The difference 
between our proposed model and the meta weight network is that, we design a simple meta mask network to learn a knowledge which can mine the pixels of corrupted labels, so as to learn a powerful segmentation network from low-quality labeled images.

\section{Meta Corrupted Pixels Mining Algorithm}
\begin{figure}[t]
    \centering
    \includegraphics[width=1.0\textwidth]{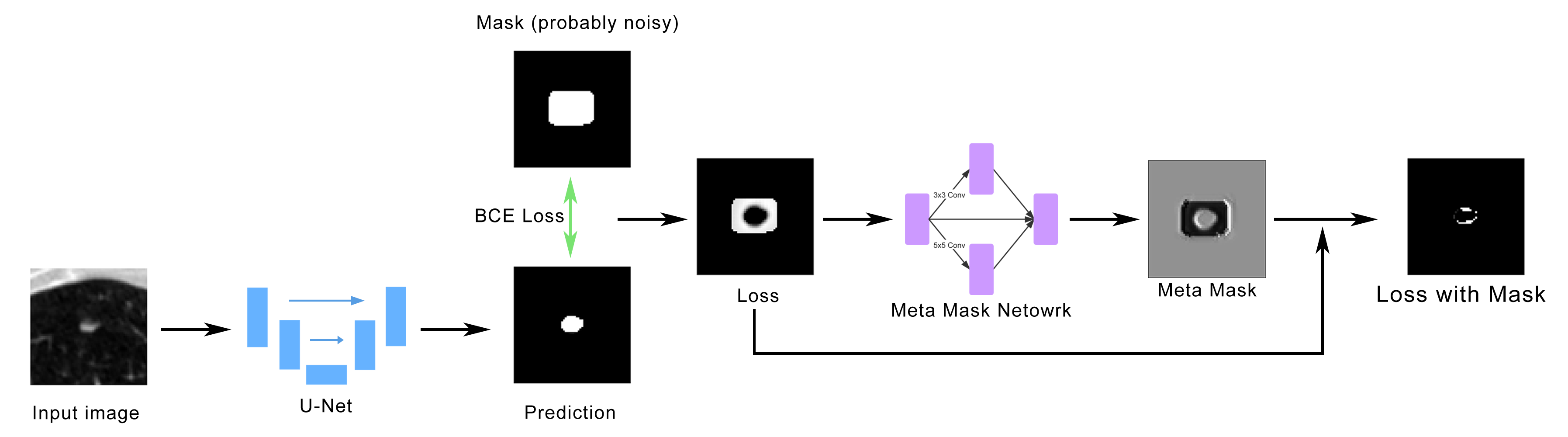}
    \caption{The architecture and workflow of one loop}
    \label{fig:architecture}
\end{figure}

We propose a novel MCPM method which can learn a powerful segmentation network from images with corrupted labels. Given a small set of images with clean labels and a large set of images with corrupted labels, our method is capable of identifying out the pixels with corrupted labels, and excluding them during the optimization procedure. As shown in Fig. \ref{fig:architecture}, our network architecture is constituted by two modules: (1) a U-Net based module for segmentation; and (2) a meta mask network for mining pixels with corrupted labels. In the following paragraphs, we will introduce our method in detail.

\subsection{Objective Functions}
Let $\mathcal{S} = \{(\mathbf{X}^{i}, \mathbf{Y}^{i})\}_{i=1}^N$ represent training images with probably noisy segmentation annotations, in which the width and height of training images are denoted by $w$ and $h$ respectively, and $N$ indicates the number of training samples. Besides, $\mathbf{Y}^{i}\in \{0,1\}^{h\times w\times c}$ denote the corrupted labels, where $c$ is the number of classes to be segmented out. First of all, we set up a segmentation network based on U-Net~\cite{unet}, which yields a pixel-level prediction $\mathbf{P}^i$ from input image $\mathbf{X}^i$. We define $\mathbf{P}^i=\mathcal{F}(\mathbf{X}^i;\mathbf{W})$ where $\mathbf{W}$ represents parameters of the segmentation network. To learn $\mathbf{W}$, an objective function is usually adopted to calculate pixel-wise loss values as function $   \mathrm{L}^i_{xy}=\textrm{loss}(\mathrm{P}^i_{xy},\mathrm{Y}^i_{xy})$, where $x\in[1,h]$ and $y\in[1,w]$ indicate the pixel coordinates. Here, the cross entropy loss function is used as the objective function.

As mentioned above, there might exist errors in segmentation annotations. These errors will severely hamper the optimization procedure, for example, providing incorrect gradient directions in the training process. A straightforward approach to cope with this issue is ignoring these pixels with incorrect labels through reweighting loss values. Inspired from~\cite{inception}, we design our meta mask network in a fully convolutional structure, which can learn an accurate mask map $\mathbf{R}^i$ for the input loss value map $\mathbf{L}^i$.    We denote $\mathbf{R}^i = \mathcal{G}(\mathbf{L}^i;\mathbf{\Theta})$, where $\mathrm{R}^i_{xy}$ indicates the reweighting factor of the pixel at $(x,y)$, and $\mathbf{\Theta}$ represents parameters of our meta mask network. Given a fixed $\mathbf{\Theta}$, the optimized solution to $\mathbf{W}$ can be found through minimizing the following objective function:
\begin{equation}\label{eq:optim-target}
    \mathbf{W}^\star(\mathbf{\Theta}) = \arg \min_{\mathbf{W}}  \frac{1}{N h w}\sum^{N}_{i=1} \sum_{x=1}^h \sum_{y=1}^w \mathrm{R}^i_{xy} \mathrm{L}^i_{xy}.
\end{equation}

To learn the parameters of our meta mask network, we introduce an additional meta dataset {\small $\hat{\mathcal{S}} = \{(\hat{\mathbf{X}}^j ,\hat{\mathbf{Y}}^j)\}_{j=1}^M$} which contains images with high-quality annotations. In particular, given an input image {\small $\hat{\mathbf{X}}^j$} and optimized parameters $\mathbf{W}^\star(\mathbf{\Theta})$, the segmentation network will produce a pixel-wise prediction map $\hat{\mathbf{P}}^j=\mathcal{F}(\hat{\mathbf{X}}^j,\mathbf{W}^\star(\mathbf{\Theta}))$ at the output layer. Again, we can obtain a loss value map $\hat{\mathbf{L}}^j$ through comparing $\hat{\mathbf{P}}^j$ against $\hat{\mathbf{Y}}^j$ according to 
the cross entropy loss function. With the optimized $\mathbf{W}$, the optimized solution to $\mathbf{\Theta}$ can be acquired through minimizing the following objective function:
\begin{equation}\label{eq:optim-meta}
    \mathbf{\Theta}^{*}(\mathbf{W}) = \arg \min_\mathbf{\Theta}  \frac{1}{M h w}\sum^{M}_{j=1}\sum^h_{x=1}\sum^w_{y=1} \hat{\mathrm{L}}^j_{xy}.
\end{equation}
In the training process, we update $\mathbf{W}$ and $\mathbf{\Theta}$ in an alternation manner. As a result, the $\mathbf{\Theta}$ can cope with the varying $\mathbf{W}$, which is beneficial to effectively mine more corrupted pixels from the predictions of the segmentation network.

\subsection{Meta Mask Network}

\begin{figure}[t]
    \centering
    \includegraphics[width=1.0\textwidth]{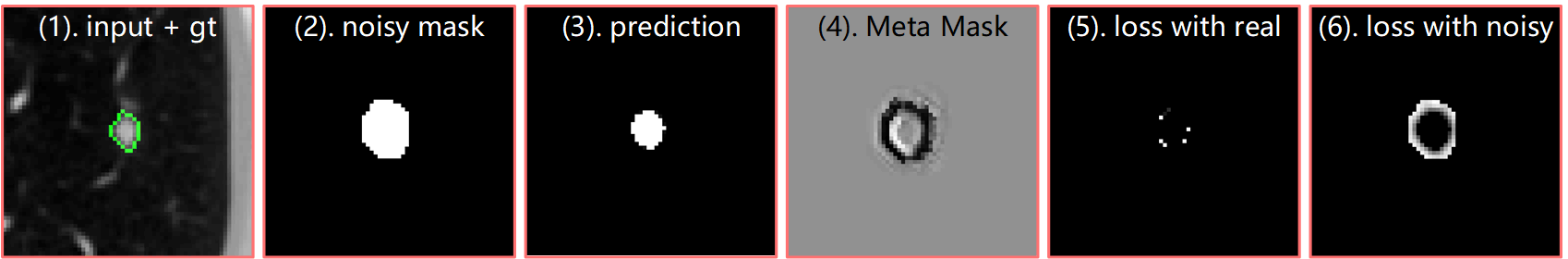}
    \caption{Illustration of how meta mask network works in the training process}
    \label{fig:fake}
\end{figure}
We take a fully convolutional structure to design our meta mask network, which can explore more local information to locate the pixels with corrupted labels. The particularities of the network are two aspects: (1) It has two convolutional layers with kernels in size of $3\times3$ and $5\times5$, which can extract multi-scale context information from $\mathbf{L}^i$. (2) The resulting outputs and input are further fused through another $1\times1$ convolutional layer, giving rise to the final mask map $\mathbf{R}^i$. This simple structure can be trained under the guidance of a few high-quality labeled samples, which will in turn help train a powerful segmentation network by using a large 
number of low-quality labeled samples.

In Fig.\ref{fig:fake}, we visualize how our meta mask network alleviates the side effect of corrupted labels in the training process, in which: (1) shows the input image and ground truth annotations; (2) indicates the corrupted labels; (3) represents the predicted result obtained by the segmentation network; (4) denotes the mined pixels of corrupted labels. As we can observe in (5) and (6), the loss between (1) and (3) is very small, while the loss between (2) and (3) is large. This indicates that our meta network can help train a powerful segmentation network with a large 
number of images accompanied with corrupted labels.

\subsection{Network Optimization}
We employ the iterative optimization algorithm to train our model. It is implemented with a single loop and mainly contains the following steps.
\begin{itemize}
\item At first, $\mathbf{W}$ and $\mathbf{\Theta}$ are randomly initialized as $\mathbf{W}^0$ and $\mathbf{\Theta}^0$.
\item For the $t$-th iteration, the parameters of the segmentation network are temporally renovated as in Eq.\eqref{eq:step_1}, via one step of gradient descent in minimizing the objective function~\eqref{eq:optim-target},
\begin{equation}
\label{eq:step_1}
\mathbf{W}'^{(t)}(\mathbf{\Theta}) = \mathbf{W}^{(t)}-\alpha \frac{1}{Nhw} \sum^{N}_{i=1}\sum^{h}_{x=1}\sum^{w}_{y=1} \mathrm{R}^{i(t)}_{xy} \frac{\partial \mathrm{L}_{xy}^i}{\partial \mathbf{W}}\bigg|_{\mathbf{W}^{(t)}},
\end{equation}
where $\alpha$ is the learning rate. $\mathrm{R}^{i(t)}_{xy}$ is computed through feeding the loss value map into the meta mask network with parameters $\mathbf{\Theta}^{(t)}$.
\item Then $\mathbf{\Theta}$ can be updated via optimizing the objective function~\eqref{eq:optim-meta},
\begin{equation}
\label{eq:step_2}
    \mathbf{\Theta}^{(t+1)} = \mathbf{\Theta}^{(t)}-\beta \frac{1}{Mhw} \sum_{j=1}^M \sum_{x=1}^h \sum_{y=1}^w \frac{\partial \hat{\mathrm{L}}^j_{xy}}{\partial \mathbf{W}'(\Theta)}\bigg|_{\mathbf{W}'^{(t)}} \frac{\partial \mathbf{W}'(\Theta)}{\partial \mathbf{\Theta}}\bigg|_{\mathbf{\Theta}^{(t)}},
\end{equation}
where $\beta$ is the learning rate.
\item Finally, $\mathbf{W}$ is updated through minimizing objective function~\eqref{eq:optim-target},
\begin{equation} \label{eq:opt-seg}
    \mathbf{W}^{(t+1)} = \mathbf{W}^{(t)} - \alpha \frac{1}{Nhw} \sum^{N}_{i=1}\sum^{h}_{x=1}\sum^{w}_{y=1} \mathrm{R}^{i(t+1)}_{xy}\frac{\partial \mathrm{L}^i_{xy}}{\partial \mathbf{W}}\bigg|_{\mathbf{W}^{(t)}}.
\end{equation}
Here $\mathrm{R}^{i(t+1)}_{xy}$ is computed through feeding the loss value map into the meta mask network with updated parameters $\mathbf{\Theta}^{(t+1)}$.
\end{itemize}

\subsection{Discussion}
Under the guidance of a small meta set with clean annotations, the meta mask network is learned in a gradient descent by gradient descent manner as shown in (\ref{eq:step_1}) and (\ref{eq:step_2}). The update of parameters in the meta mask network is dependent on the gradients of losses calculated on pixels from both meta and training images. After putting (\ref{eq:step_1}) into (\ref{eq:step_2}), it can be easily observed that the ascending direction of the weight coefficient of every pixel relies on the inner product (it can also be interpreted as a similarity metric) between the gradient of the pixel (formulated as $\frac{\partial \textrm{L}_{xy}^i}{\partial \mathbf W}|_{\mathbf W^{(t)}}$) and the average gradient of pixels of meta images (formulated as $\frac{1}{Mhw} \sum_{j=1}^M \sum_{x=1}^h \sum_{y=1}^w \frac{\partial \hat{\textrm L}^j_{xy}}{\partial \mathbf W'(\Theta)}|_{\mathbf W'^{(t)}}$). A positive inner product pushes the parameters of the meta mask network towards a direction which can give rise to a larger weighting coefficient for the corresponding pixel; a negative inner product pushes the network towards the opposite direction. This is the reason why our method can effectively identify corrupted pixels.

\section{Experiments}
\subsection{Datasets and Metrics} \label{sec:dataset}
Two datasets are exploited to validate the superiority of our method in medical image segmentation with noisy annotations, including LIDC-IDRI (Lung Image Database Consortium and Image Database Resource Initiative) \cite{lidc,lidc-data,tcia}  and LiTS (Liver Tumor Segmentation) \cite{LiTS} . $64\times64$ patches covering lesions are cropped out as training or testing samples.
\begin{itemize}
\item[(1)] The LIDC-IDRI dataset contains 1018 lung CT scans from 1010 patients with lesion masks annotated by four experts. 3591 patches are cropped out. They are split into a training set of 1906 images and a testing set of 1385 images. The remain 300 images are used as the meta set. 
\item[(2)] The LiTS dataset includes 130 abdomen CT scans accompanied with annotations of liver tumors. 2214 samples are sampled from this dataset. 1471, 300 and 443 images are used for training, meta weight learning, and testing respectively.
\end{itemize}
Three metrics, including IOU(also referred as the Jaccard Index), Dice coefficient and Hausdorff distance, are employed for quantitatively measuring performances of segmentation algorithms.

\subsubsection{Synthesizing Noisy Annotations.} In practice, it is difficult to localize the boundary of the target region during the annotating procedure. Considering this phenomenon, we synthesize noisy annotations through creating masks which loosely encompasses target lesions. We use 2 operators to simulate corrupted annotations. 1) The dilation morphology operator is employed to extend the foreground region by several pixels (randomly drawn from $[0,6]$). 2) The toolkit of deformation provided in ElasticDeform\cite{elsticdeform,unet,Cciccek_Abdulkadir_Lienkamp:2016}, which includes more complicate operations such as rotation, translation, deformation and morphology dilation, is used to contaminate ground-truths of training images. In our experiment, only a part of samples are contaminated with the above strategies. We denote the percent of images which are selected out to generate noisy labels as $r$.

\subsection{Implementation Details}
Adam and SGD is used to optimize to network parameters on LIDC-IDRI and LiTS, respectively. The learning rates $\alpha$ and $\beta$ are initialized as $10^{-4}$ and $10^{-3}$ respectively, and decayed by 0.1 in $20^{th}$ epoch and $40^{th}$ epoch. The batch size is set as 128. All models are trained with 120 epochs.

\begin{figure}[t]
    \centering
    \includegraphics[width=0.9\textwidth]{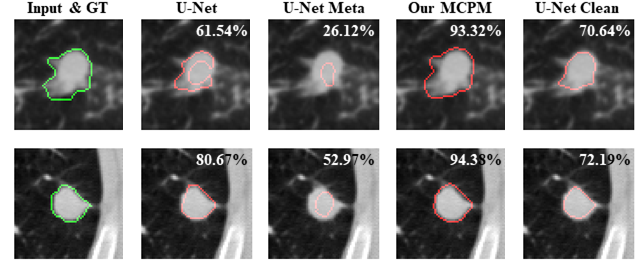}
    \caption{Visualization of segmentation results. Green and red contours indicate the ground-truths and segmentation results, respectively. The Dice value is shown at the top-right corner, and our method produces much better results than other methods. }
    \label{fig:exper}
\end{figure}
\subsection{Experimental Results}
\subsubsection{Comparison with Existing Methods.} Without specification, $r$ is set to $40\%$ in experiments of this section which means annotations of 40\% training images are contaminated. We compare our method against 7 existing segmentation models which are proposed to deal with ambiguous, low-quality or insufficient annotations on the LIDC-IDRI dataset: \textbf{Prob U-Net} \cite{probunet}, \textbf{Phi-Seg} \cite{phiseg},  \textbf{UA-MT} \cite{uamt} modified for 2D segmentation, \textbf{Curriculum} \cite{curriculum}, \textbf{Few-Shot GAN} \cite{few}, \textbf{Quality Control} \cite{vaeunet}, \textbf{$U^{2}$ Net} \cite{huang20193d},  and \textbf{MWNet}\cite{meta_weight_learning} which is integrated with U-Net. All above models and the baseline U-Net are trained with mixed images of the training set and the meta set. We also implement another variant of U-Net which is trained merely using images from the meta set (indicated by `U-Net Meta'). We also report the result of U-Net trained using images with clean labels (indicated by `U-Net Clean'). As shown in Table.~\ref{table:12_performance}.Our method performs significantly better than other methods. For example, the Dice value of our method surpasses that of the second best method MWNet by 3.4\%. Additionally, our method outperforms baseline U-Net models by a large margin. It even achieves promising performance which is comparable to the result of `U-Net Clean'. This indicates that the impact of incorrect annotations fabricated as in Section \ref{sec:dataset} is almost eliminated. Visualization examples are shown in Fig. \ref{fig:exper}.

\begin{table*}[t]
    \caption{Results of segmentation models on LIDC-IDRI. 
    }
    \label{table:12_performance}
    \centering
	\begin{scriptsize}
	\setlength{\tabcolsep}{1.5mm}{

        \begin{tabular}{p{2.46cm}<{\centering}|p{1.06cm}<{\centering}|p{1.06cm}<{\centering}|p{1.5cm}<{\centering}|p{1.06cm}<{\centering}|p{1.06cm}<{\centering}|p{1.5cm}<{\centering}}
			\toprule
			Noisy & \multicolumn{3}{c|}{Dilation} & \multicolumn{3}{c}{ElasticDeform}\\
			\hline
			Model name & mIOU & Dice & Hausdorff & mIOU & Dice & Hausdorff \\
			\hline
            U-Net&62.53&75.56&1.9910&65.01&76.17&1.9169 \\
            U-Net Meta&60.91&72.21&2.0047&60.91&72.21&2.0047 \\
            Prob U-Net&66.42&78.39&1.8817&68.43&79.50&1.8757 \\
            Phi-Seg&67.01&79.06&1.8658&68.55&81.76&1.8429 \\
            UA-MT&68.18&80.98&1.8574&68.84&82.47&1.8523 \\
            Curriculum&67.78&79.54&1.8977&68.18&81.30&1.8691 \\
            Few-Shot GAN&67.74&78.11&1.9137&67.93&77.83&1.9223 \\
            Quality Control&65.00&76.50&1.9501&68.07&77.68&1.9370 \\
            $U^{2}$ Net&65.92&76.01&1.9666&67.20&77.05&1.9541 \\
            MWNet&71.56&81.17&1.7762&71.89&81.04&1.7680 \\
            Our MCPM& \textbf{74.69}&\textbf{84.64}&\textbf{1.7198}&\textbf{75.79}&\textbf{84.99}&\textbf{1.7053} \\
            U-Net Clean&\underline{75.73}&\underline{83.91}&\underline{1.7051}&\underline{75.73}&\underline{83.91}&\underline{1.7051} \\
			\bottomrule
        \end{tabular}}
    
    \end{scriptsize}
\end{table*}

\subsubsection{Results with Various $r$-s.} In this section, we vary the percent of noisy images $r$ from 0 to 0.8. The segmentation results of four methods on LIDC-IDRI and LiTS are presented in Table \ref{table:inner_performance}. On the LIDC-IDRI dataset, our method performs better than other methods when noises are introduced into the training set. On the LiTS dataset, our method exceeds other methods consistently under all cases.   

\begin{table}[t]
	\caption{Results~(mIOU) of segmentation models using various $r$-s.}
	\label{table:inner_performance}
	\centering
	\begin{scriptsize}
		\setlength{\tabcolsep}{1.6mm}{
		\begin{tabular}{p{1.95cm}<{\centering}|p{0.65cm}<{\centering}|p{0.65cm}<{\centering}|p{0.65cm}<{\centering}|p{0.65cm}<{\centering}|p{0.65cm}<{\centering}|p{0.65cm}<{\centering}|p{0.65cm}<{\centering}|p{0.65cm}<{\centering}|p{0.65cm}<{\centering}|p{0.65cm}<{\centering}}
			\toprule
			Dataset Name & \multicolumn{5}{c|}{LIDC-IDRI} & \multicolumn{5}{c}{LiTS}\\
			\hline
			$r$ & 0.8 & 0.6 & 0.4 & 0.2 & 0 & 0.8 & 0.6 & 0.4 & 0.2 & 0 \\
			\hline
			U-Net & 42.64& 51.23& 62.53& 69.88&75.73& 37.18&43.55& 46.41&51.20&61.07\\
			Prob U-Net & 52.13& 60.81& 66.42& 71.03& \textbf{76.37}& 40.16&45.90& 49.22&53.97&60.60\\
			MWNet & 61.28& 67.33& 71.56& 72.07& 74.40& 43.14&44.97& 51.96&58.65&59.18\\
			Our MCPM & \textbf{67.60}& \textbf{68.97}& \textbf{74.69}& \textbf{74.87}& 75.26& \textbf{45.09}&\textbf{48.76}& \textbf{55.17}&\textbf{62.04}&\textbf{62.68}\\
			\bottomrule
		\end{tabular}}
	\end{scriptsize} 
\end{table}


\section{Conclusion}
We proposed a novel Meta Corrupted Pixels Mining method to alleviate the side effect of corrupted label in medical image segmentation. Given a small number of high-quality labeled images, the deduced learning regime make our meta mask network able to locate the pixels having corrupted labels, which can be used to help train a powerful segmentation network from a large number of low-quality labeled images. Extensive experiments on two datasets, LIDC-IDRI and LiTS, show that the proposed method can achieve the state-of-the-art performance in medical image segmentation.

\noindent \textbf{Acknowledgments}. This work is jointly supported by the National Key Research and Development Program of China under Grant No. 2017YFA0700800, the National Natural Science Foundation of China Grant No. 61629301, 61976171, and the Key Research and Development Program of Shaanxi Province of China under Grant No. 2020GXLH-Y-008.

\bibliographystyle{splncs04}
\bibliography{paper175}
\end{document}